\documentclass[runningheads]{llncs}
\usepackage{graphicx}
\usepackage{hyperref}
\usepackage{tikz}
\def\checkmark{\tikz\fill[scale=0.4](0,.35) -- (.25,0) -- (1,.7) -- (.25,.15) -- cycle;}

\begin{document}
\title{TransDocAnalyser: A Framework for Offline Semi-structured Handwritten Document Analysis in the Legal Domain}
\titlerunning{TransDocAnalyser:  A framework for semi-structured legal document analysis}

\author{{Sagar Chakraborty\inst{1,2} \and
Gaurav Harit\inst{2} \and
Saptarshi Ghosh\inst{3}}}
\authorrunning{{Chakraborty, Sagar et al.}}

\institute{Wipro Limited, Salt Lake, Kolkata, India \and
Department of Computer Science and Engineering, Indian Institute of Technology, Jodhpur, Rajasthan, India
\newline
\email{ chakraborty.4@iitj.ac.in, gharit@iitj.ac.in} \and
Department of Computer Science and Engineering, Indian Institute of Technology, Kharagpur, West Bengal, India\\
\email{ saptarshi@cse.iitkgp.ac.in}}
\maketitle % typeset the header of the contribution
\begin{abstract}
State-of-the-art offline Optical
Character Recognition (OCR) frameworks perform poorly on semi-structured handwritten domain-specific documents due to their inability to localize and label form fields with domain-specific semantics. Existing techniques for semi-structured document analysis have primarily used datasets comprising invoices, purchase orders, receipts, and identity-card documents for benchmarking. 
In this work, we build the first semi-structured document analysis dataset in the legal domain by collecting a large number of First Information Report (FIR) documents from several police stations in India. This dataset, which we call the FIR dataset, is more challenging than most existing document analysis datasets, since it combines a wide variety of handwritten text with printed text.
We also propose an end-to-end framework for offline processing of handwritten semi-structured documents, and benchmark it on our novel FIR dataset. 
Our framework used Encoder-Decoder architecture for localizing and labelling the form fields and for
recognizing the handwritten content. 
The encoder consists of Faster-RCNN and Vision Transformers. Further the Transformer-based decoder architecture is trained with a domain-specific tokenizer. We also propose a post-correction method to handle recognition errors pertaining to the domain-specific terms.  
Our proposed framework achieves state-of-the-art results on the FIR dataset outperforming several existing models.

\end{abstract}

\section{Introduction}
\begin{figure}[tb]
\includegraphics[width=12cm, height=10cm]{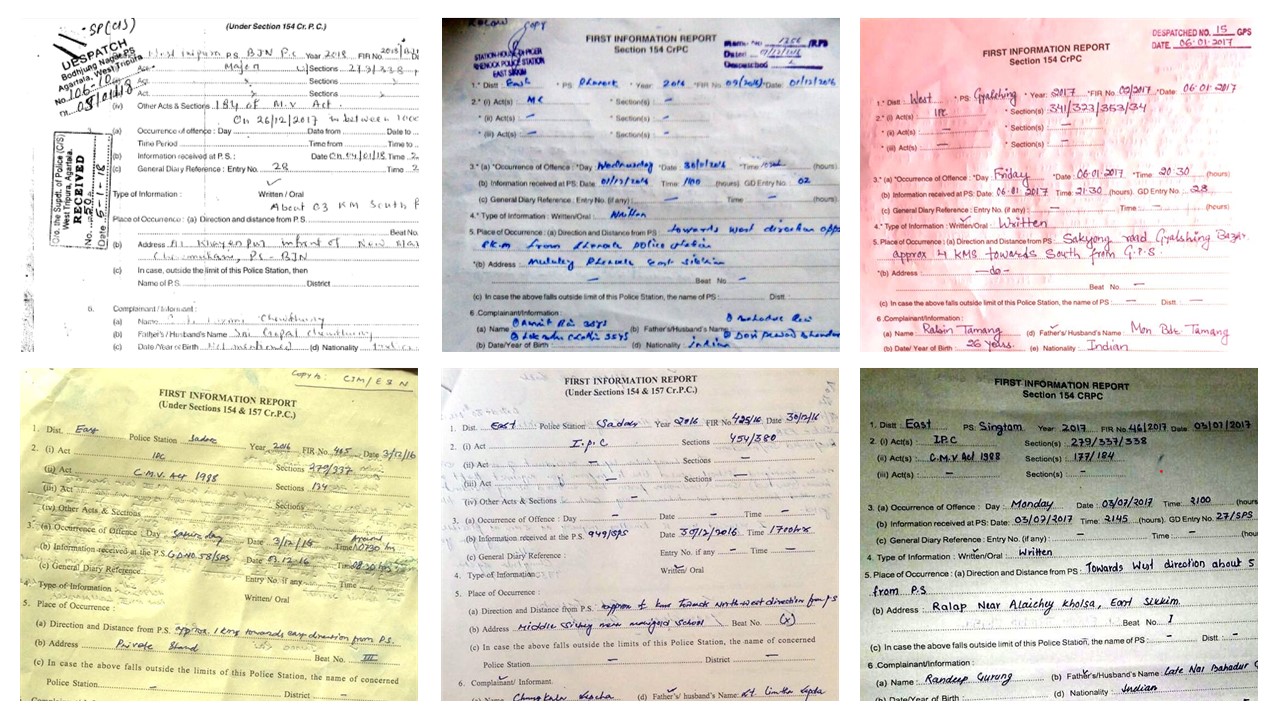}
\caption{Examples of First Information Report (FIR) documents from different police stations in India. The FIR dataset developed in this paper consists of a wide variety of such semi-structured FIR documents containing both printed and handwritten text. 
} 
\label{fig:fir-examples}
\end{figure}

Semi-Structured documents are widely used in many different industries. Recent advancement in digitization has increased  the demand for analysis of scanned or mobile-captured semi-structured documents. 
Many recent works have used different deep learning techniques to solve some of the critical problems in processing and layout analysis of semi-structured documents~\cite{survery2020dlocr,invoiceocr,donut}. 
Semi-structured documents consist of printed, handwritten, or hybrid (both printed and handwritten) text forms. In particular, hybrid documents (see Figure~\ref{fig:fir-examples}) are more complex to analyze since they require segregation of printed and handwritten text and subsequent recognition. 
With recent advancements, the OCR accuracy has improved for printed text; however, recognition of handwritten characters is still a challenge due to variations in writing style and layout.

Earlier works have focused on techniques for  layout analysis, named-entity recognition, offline handwriting recognition, etc., but sufficient work has \textit{not} been done on developing an  end-to-end framework for processing semi-structured documents. 
A general end-to-end framework can be easily fine-tuned for domain-specific requirements. In this paper we present the first framework for semi-structured document analysis applied to legal documents.

There have been many works on legal documents, such as on case document summarization~\cite{bhattacharya2019comparative}, 
%estimating the similarity between legal case documents~\cite{docsim_bhattacharya20}, 
relevant statute identification from legal facts~\cite{paul2022lesicin}, pretraining language models on legal text~\cite{inlegalbert} and so on. 
But almost all prior research in the legal domain has focused on textual data, and {\it not} on document images. 
In particular, the challenges involved in document processing and layout analysis of legal documents is unattended, even though these tasks have become important due to the increasing availability of scanned/photographed legal documents.

%In 2020, Indian Supreme Court decided to digitized all the documents related to legal cases and make available in digital library for people and practitioners.

In this work, we build the first dataset for semi-structured document analysis in the legal domain.
To this end, we focus on \textbf{First Information Report} (FIR) documents from India.
An FIR is usually prepared by police stations in some South Asian countries when they first get a complaint by the victim of a crime (or someone on behalf of the victim).\footnote{\url{https://en.wikipedia.org/wiki/First_information_report}} %\cite{wiki_fir}
An FIR usually contains a lot of details such as the date, time, place, and details of the incident, the names of the person(s) involved, a list of the statutes (written laws, e.g., those set by the Constitution of a country) that might have been violated by the incident, and so on.
The FIRs are usually written on a printed form, where the fields are filled in by hand by police officials (see examples in Figure~\ref{fig:fir-examples}).
%In 2021, a total of 60,96,310 crimes were reported through First Information Report(FIR) in India alone\cite{ref_conf34}. 
It is estimated that more than 6 million FIRs are filed every year across thousands of police stations in various states in India.
Such high volumes lead to inconsistent practices in-terms of handwriting, layout structure, scanning procedure, scan quality, etc., and introduce huge noise in the digital copies of these documents. 
%Classical image processing techniques and deep learning based models like convolutional neural networks, encoder-decoder based architectures, generative adversarial networks etc have been used to address these issues. 
%More recent works have also shown explicit image processing stages can be bypassed in many architectures\cite{ref_conf8}. 

Our target fields of interest while processing FIR documents are the handwritten entries (e.g., name of the complainant, the statutes violated) which are challenging to identify due to the wide variation in handwriting. 
To form the dataset, which we call the {\bf FIR dataset}, we created the meta-data for the target fields by collecting the actual text values from the police databases, and also annotated the documents with layout positions of the target fields.
The FIR dataset is made publicly available at \url{https://github.com/LegalDocumentProcessing/FIR_Dataset_ICDAR2023}.

The FIR dataset is particularly challenging since its documents are of mixed type, with both printed and handwritten text. 
Traditional OCR identifies blocks of text strings in documents and recognizes the text from images by parsing from left to right~\cite{thomasocr}. 
NLP techniques like named-entity recognition (NER), which uses raw text to find the target fields, cannot be applied easily, since traditional OCRs do not work well in recognition of mixed documents with handwritten and printed characters occurring together.
Another drawback of traditional OCRs in this context is their inability to recognise domain-specific words due to their general language-based vocabulary. 
%For instance, Indian legal documents are full of terms that are not English, such as ‘Tehsildar’ (Collector), ’Daroga’ (Constable), and so on. Understanding these terms can be crucial depending on the particular case and the intended application\cite{ref_conf10}. 
In this work, we propose a novel framework for analysing such domain-specific semi-structured documents. 
The contributions of the proposed framework as follows:
\begin{enumerate}

\item We use a FastRCNN + Vision Transformer-based encoder trained for target field localization and classification. 
%We do not need a two-stage process of applying and name-entity recognition to deal with challenging handwritten content in the target field.
We also deploy a BERT-based text decoder that is fine-tuned to incorporate legal domain-specific vocabulary.

\item We use a domain-specific pretrained language model~\cite{inlegalbert} to improve the recognition of domain-specific text (legal statutes, Indian names, etc.).  This idea of using a domain-specific language model along with OCR is novel, and has a wider applicability over other domains (e.g., finance, healthcare, etc) where this technique can be used to achieve improved recognition from domain-specific documents.

\item We improve the character error rate (CER) by reducing the ambiguities in OCR through a novel domain-specific post-correction step. Using domain knowledge, we created a database for each target field (such as Indian names, Indian statutes, etc.) to replace the ambiguous words from OCR having low confidence using a combination of TF-IDF vectorizer and K-Nearest Neighbour classifier. 
This novel post-correction method to handle recognition errors pertaining to proper nouns, enables our proposed framework to outperform state-of-the-art OCR models by large margins. 

\end{enumerate}
To summarize, in this work we build the first legal domain-specific dataset for semi-structured document analysis. 
We also develop a framework to localise the handwritten target fields, and fine-tune a transformer-based OCR (TrOCR) to extract handwritten text. We further develop post-correction techniques to improve the character error rate. 
%We benchmark the performance of our framework on our novel FIR dataset.
To our knowledge, the combination of Faster-RCNN and TrOCR with other components, such as Vision Transformer and legal domain-specific tokenizers, to create an end-to-end framework for processing offline handwritten semi-structured documents is novel, and can be useful for analysis of similar documents in other domains as well.

\section{Related Work}

We briefly survey four types of prior works related to our work -- 
(i)~related datasets, (ii)~works addressing target field localization and classification, (iii)~handwritten character recognition, and (iv)~works on post-OCR correction methods 

\vspace{2mm}
\noindent \textbf{Related Datasets:} There exist several popular datasets for semi-structured document analysis.
FUNSD~\cite{funsd} is a very popular dataset for information extraction and layout analysis. 
FUNSD dataset is a subset of RVL-CDIP dataset~\cite{harley_imageclassification}, and contains 199 annotated financial forms.
The SROIE dataset~\cite{icdar19_dataset} contains 1,000 annotated receipts having 4 different entities, and is used for receipt recognition and information extraction tasks.
The CloudSCan Invoice dataset~\cite{cloudscan} is a custom dataset for invoice information extraction. The dataset contained 8 entities in printed text.

Note that no such dataset exists in the legal domain, and our FIR dataset is the first of its kind. Also, the existing datasets contain only printed text, while the dataset we build contains a mixture of printed  and hand-written text (see Table~\ref{tab:dataset-compare} for a detailed comparison of the various datasets).

%\subsubsection{Localization and Labelling of field components}
\vspace{2mm}
\noindent \textbf{Localization and Labelling of field components:}
Rule-based information extraction methods (such as the method developed by Kempf et al.~\cite{constum_2022} and many other methods) could be useful when documents are of high quality and do not contain handwritten characters. But when document layouts involve huge variations, noise and handwritten characters, keyword-based approaches fail to provide good results.
Template-based approaches also fail due to scanning errors and layout variability~\cite{watanabe_tableform,graph_form_analysis,amano_tableform}.

Srivastava et al.~\cite{divya_form} developed a graph-based deep network for predicting the associations between field labels and field values in handwritten form images. They considered forms in which the field label comprises printed text and field value can be handwritten text; this is similar to what we have in the FIR dataset developed in this work. 
To perform association between the target field
labels and  values, they formed a graphical representation of the textual scripts using their associated layout position. 

In this work, we tried to remove the dependency on OCR of previous works~\cite{divya_form} by using layout information of images to learn the positions of target fields and extract the image patches using state-of-the-art object detection models such as~\cite{ddetr,efficient_det,faster_rcnn}. 

Zhu et. al. \cite{ddetr} proposed attention modules that only attend to a small set of key sampling points around a reference, which can achieve better performance than baseline model~\cite{detr} with 10$\times$ less training epochs. Tan et. al.~\cite{efficient_det} used weighted bi-directional feature pyramid network (BiFPN), which allows easy and fast multi-scale feature fusion. Ren et al~\cite{faster_rcnn} proposed an improved version of their earlier work~\cite{fast_rcnn} provides comparative performances with~\cite{ddetr,efficient_det} with lower latency and computational resources on FIR dataset. Hence, we use Faster RCNN model in this framework for localization and classification of the field component.
%Faster RCNN \cite{ref_conf38} which is an improved version of Fast-RCNN \cite{ref_conf47} which we have used in this work for localization and classification of the field component.
%The image are passed into Faster RCNN~\cite{faster_rcnn} feeding  directly into the OCR to generate the targeted image patches; this approach improves the quality of handwritten character recognition. 

%\subsubsection{Handwritten Character Recognition}
\vspace{2mm}
\noindent \textbf{Handwritten Character Recognition:}
Offline handwriting recognition has been a long standing research interest. 
The works~\cite{soumen_thinning,soumen_contour,soumen_topo} presented novel features based on structural features of the strokes and their spatial relations with a character, as visible from different viewing directions on a 2D plane. Diesendruck et al.~\cite{Diesendruck_htr} used Word Spotting to directly recognise handwritten text from images. 
%A more recent work~\cite{hiteshi_siamese}  proposed a SiameseCNN Network that learns to discriminate features, which can then be used for identification of a pair of images  containing similar or dissimilar characteristics. 
The conventional text recognition task is usually framed as an encoder-decoder problem where the traditional methods\cite{thomasocr} leveraged CNN-based~\cite{cnn_imagenet} encoder for image understanding and LSTM-based~\cite{lstm_1997} decoder for text recognition. 

Chowdhury et al.~\cite{chowdhury2018efficient} combined a deep convolutional network with a recurrent Encoder-Decoder network to map an image to a sequence of characters corresponding to the text present in the image. 
%Recently the Transformer has gained popularity in the field of natural language processing. 
Michael, Johannes et al. \cite{stos_htr} proposed a sequence-to-sequence model combining a convolutional neural network (as a generic feature extractor) with a recurrent neural network to encode both the visual information, as well as the temporal context between characters in the input image. 
Further, Li et al.~\cite{trocr} used for the first time an end-to-end Transformer-based encoder-decoder OCR model for handwritten text recognition and achieved SOTA results. The model~\cite{trocr} is convolution-free unlike previous methods, and does not rely on any complex pre/post-processing steps. The present work leverages this work and extends its application in legal domain.

%\subsubsection{Post OCR correction}
\vspace{2mm}
\noindent \textbf{Post-OCR correction:}
Rectification of errors in the recognised text from the OCR would require extensive training which is computation heavy. Further, post-OCR error correction requires a large amount of annotated data which may not always be available. After the introduction of the Attention mechanism and BERT model, many works have been done to improve the results of the OCR using language model based post-correction techniques. However, Neural Machine Translation based approaches as used by Duong et al. \cite{duong_postocr} are not useful in the case of form text due to the lack of adequate context and neighbouring words. We extend the idea used in the work of Trstenjak et al. \cite{knn_tfidf} where they used edit distance and cosine similarity to find the matching words. In this paper we used K-nearest neighbour with edit distance to find best matches for the words predicted with low confidence score by the OCR.

\section{The FIR Dataset} \label{sec:dataset}

First Information Report (FIR) documents contain  details about  incidents of cognisable offence, that are written at police stations based on a complaint.
FIRs are usually filed by a police official filling up a printed form; hence the documents contain both printed and handwritten text. 
In this work, we focus on FIR documents written at police stations in India.
Though the FIR forms used across different Indian states mostly have a common set of fields, there are some differences in their layout (see examples in Fig.~\ref{fig:fir-examples}). 
To diversify the dataset, we included FIR documents from the databases of various police stations across several Indian states -- West Bengal\footnote{\url{http://bidhannagarcitypolice.gov.in/fir_record.php}}, Rajasthan\footnote{\url{https://home.rajasthan.gov.in/content/homeportal/en.html}}, Sikkim\footnote{\url{https://police.sikkim.gov.in/visitor/fir}}, Tripura\footnote{\url{https://tripurapolice.gov.in/west/fir-copies}} and Nagaland\footnote{\url{https://police.nagaland.gov.in/fir-2/}}. 

%\cite{bidhannagar_fir,rajasthan_police,tripura_police,naga_police,sikkim_police}

As stated earlier, an FIR contains many fields including the name of the complainant, names of suspected/alleged persons, statutes that may have been violated, date and location of the incident, and so on.
In this work, we selected \textit{four target fields} from FIR documents for the data annotation and recognition task -- 
(1)~\textit{Year} (the year in which the complaint is being recorded), 
(2)~\textit{Complainant's name} (name of the person who lodged the complaint), 
(3)~\textit{Police Station} (name of the police station that is responsible for investigating the particular incident), and
(4)~\textit{Statutes} (Indian laws that have potentially been violated in the reported incident; these laws give a good indication of the type of the crime).
We selected these four target fields because  we were able to collect the gold standard for these four fields from some of the police databases.
%one of the police commissionerate databases \cite{bidhannagar_fir} 
Also, digitizing these four fields would enable various societal analysis, such as analysis of the nature of crimes in different police stations, temporal variations in crimes, and so on.

%We used this gold standard to train our OCR and object detection models. 

\begin{figure}
\includegraphics[width=12cm, height=8cm]{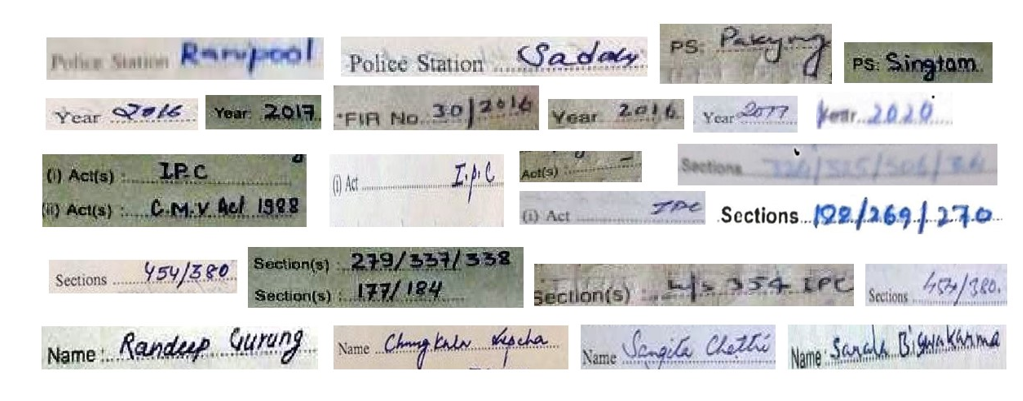}
\vspace{-3mm}
\caption{Sample of various entities present in First Information Reports with different writing styles, distortions and scales.} 
\label{fig:sample-entities}
\vspace{-3mm}
\end{figure}

\vspace{2mm}
\noindent \textbf{Annotations:}
We manually analysed more than 1,300 FIR documents belonging to different states, regions, police stations, etc. We found that FIR documents from the same region / police station tend to have the similar layout and form structure. 
Hence we selected a subset of 375 FIR documents with reasonably varying layouts / form structure, so that this subset covers most of the different variations. These 375 documents were manually annotated.
Annotations were done on these documents using LabelMe annotation tool\footnote{\url{https://github.com/wkentaro/labelme}} to mark the bounding boxes of the target fields.  

\begin{figure}
\centering
\includegraphics[width=8cm, height=6cm]{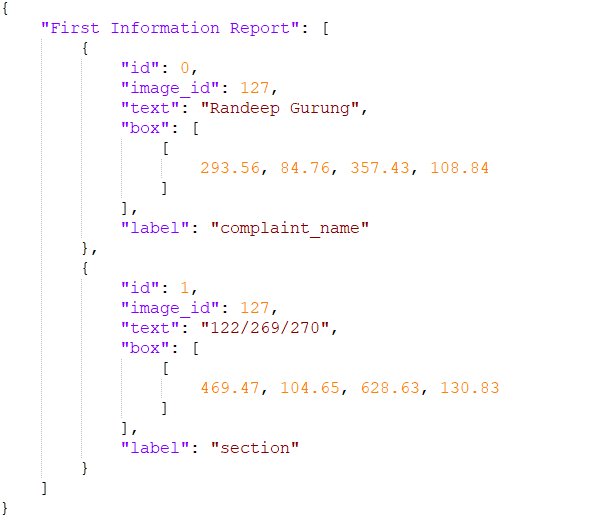}
\caption{Examples of ground truth annotations for two of the entities shown in Figure~\ref{fig:sample-entities}} 
\label{fig:ground-truth}
\end{figure}

Figure~\ref{fig:sample-entities} shows some samples of various entities present in our dataset, and Figure~\ref{fig:ground-truth} shows examples of ground truth annotations for two of the entities in Figure~\ref{fig:sample-entities}.
In the ground truth, each bounding box has four co-ordinates (X\_left, X\_width, Y\_right, Y\_height) which describe the position of the rectangle containing the field value for each target field.

\vspace{2mm}
\noindent \textbf{Train-test split:}
During the annotation of our dataset, we identified 79 different types of large scale variations, layout distortions/deformations, which we split into training and testing sets. 
We divided our dataset (of 375 document images) such that 300 images are included in the training set and the other 75 images are used as the test set. During training, we used 30\% of training dataset as a validation set.
Table~\ref{tab:dataset-stats} shows the bifurcation statistics for training and test sets.
%The annotators annotated 1,537 labels which contain 2,287 words in total.

\begin{table}[t]
\caption{FIR Dataset statistics}
\label{tab:dataset-stats}
\centering
\setlength{\tabcolsep}{4pt}
\begin{tabular}{l|c|c|c|c}
\hline
Split & Images & Layout & Words & Labels \\
\hline
Training & 300 & 61 & 1,830 & 1,230 \\
Testing & 75 & 18 & 457 & 307 \\
\hline
\end{tabular}
\vspace{-5mm}
\end{table}

\vspace{2mm}
\noindent \textbf{Preprocessing the images:}
For Faster-RCNN we resized the document images to a size of 1180 $\times$ 740, and used the bounding boxes and label names to train the model to predict and classify the bounding boxes. 
We convert the dataset into IAM Dataset format~\cite{iam_dataset} to fine-tune the transformer OCR.

\vspace{2mm}
\noindent \textbf{Novelty of the FIR dataset:}
We compare our FIR dataset\footnote{ \url{https://github.com/LegalDocumentProcessing/FIR_Dataset_ICDAR2023}} with other datasets for semi-structure document analysis in Table~\ref{tab:dataset-compare}. 
The FIR dataset contains both printed and handwritten information which makes it unique and complex compared to several other datasets. 
Additionally, the FIR dataset is the first dataset for semi-structured document analysis in the legal domain.

\begin{table}[tb] 
\caption{Comparison of the FIR dataset with other similar datasets}
\label{tab:dataset-compare}
\centering
\begin{tabular}{|l|l|r|l|l|l|}
\hline
Dataset & Category & \#Images & \multicolumn{2}{|l|}{Text Type} & \#Entites \\
\cline { 4 - 5 }
 &  &  & \textbf{Printed} & \textbf{Handwritten} & \\
\hline
FUNSD \cite{funsd} & Form & 199 & \checkmark & x & 4 \\
\hline
SROIE \cite{icdar19_dataset} & Receipt & 1000 & \checkmark & x & 4 \\

\hline
Cloud Invoice \cite{cloudscan} & Invoice & 326571 & \checkmark & x & 8 \\
\hline
FIR ({\bf Ours}) & Form & 375 & \checkmark & \checkmark & 4 \\
\hline
\end{tabular}
\vspace{-5mm}
\end{table}

\section{The TransDocAnalyser Framework}

We now present TransDocAnalyser, a framework for offline processing of handwritten semi-structured documents, by adopting Faster-RCNN and Transformer-based encoder-decoder architecture, with post-correction to improve performance. 

\subsection{The Faster-RCNN architecture}

Faster-RCNN~\cite{faster_rcnn} is a popular object detection algorithm that has been adopted in many real-world applications.
It builds upon the earlier R-CNN~\cite{orig_rcnn} and Fast R-CNN~\cite{faster_rcnn} architectures. 
We pass the input images through the Faster-RCNN network to get the  domain-specific field associations and extract the image patches from the documents.

%Note that, along with Faster-RCNN, we tried multiple state-of-the-art object detection models, such as Deformable Transformer~\cite{ddetr}and EfficientDet~\cite{efficientdet}. An internal evaluation over the FIR dataset showed that the performances of all these segmentation models are comparable in terms of accuracy, but we observed a high latency during the inference in case of the other models (compared to Faster-RCNN). Hence we chose to use Faster-RCNN which is equally accurate and faster compared to other models on the proposed FIR dataset. 

\begin{figure}[tb]
\centering
\includegraphics[width=\textwidth, height=5.6cm]{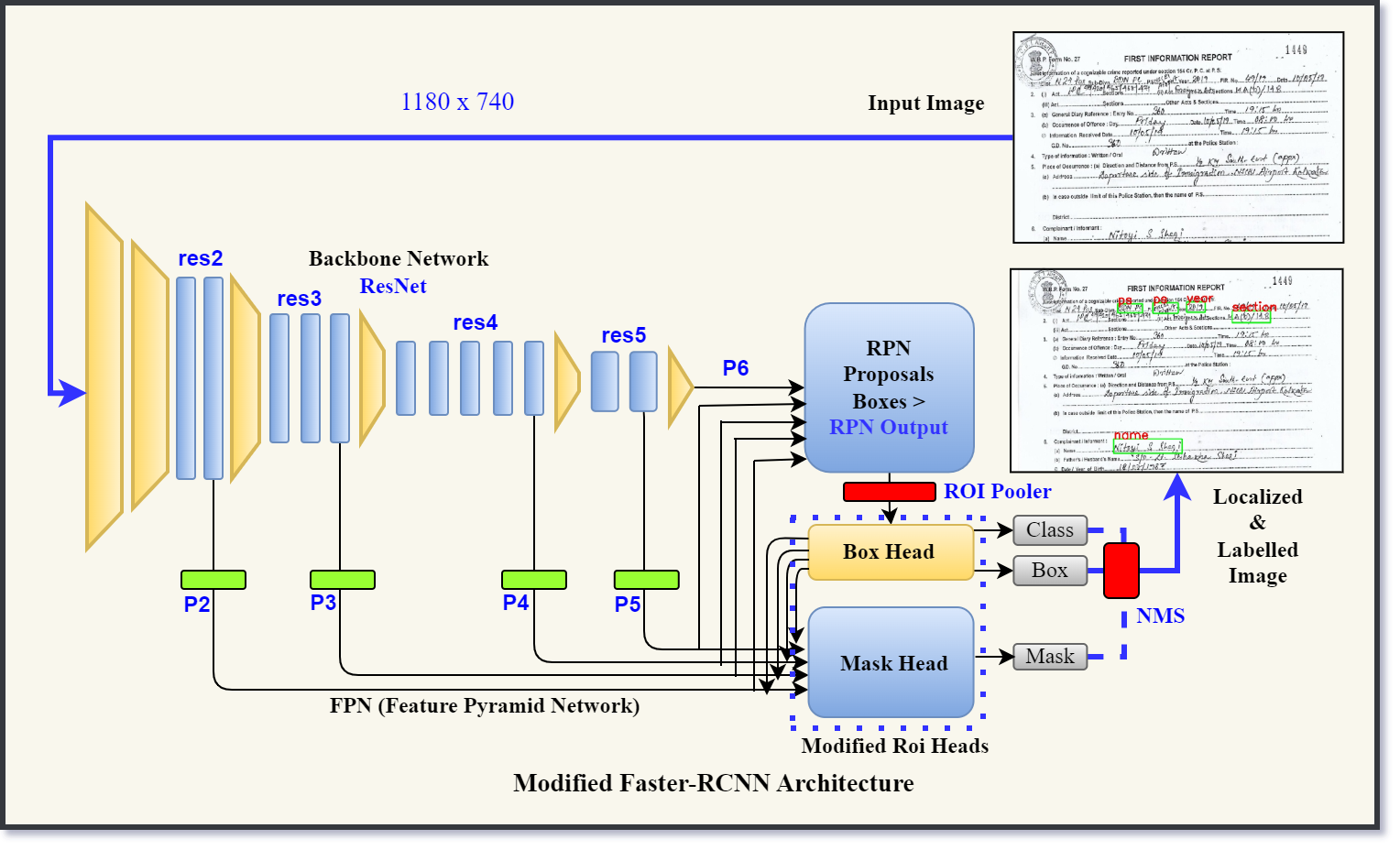}
\caption{Modified Faster-RCNN based architecture for target field localization and labelling} 
\label{fig:faster-rcnn-arch}
\vspace{-5mm}
\end{figure}

Our modified Faster-RCNN architecture consists of three main components (as schematically shown in Figure~\ref{fig:faster-rcnn-arch})-- 
(1)~Backbone Network , (2)~Region Proposal Network (RPN), and (3)~ROI Heads as detailed below.

\vspace{1mm}
\noindent \textbf{(1) Backbone Network:} ResNet-based backbone network is used to extract multi-scaled feature maps from the input -- that are named as P2, P3, P4 , P8 and so on -- which are scaled as 1/4th, 1/8th, 1/16th and so on. 
This backbone network is FPN-based (Feature Pyramid network)~\cite{lin_fpn} which is multi-scale object detector invariant to the object size.

\vspace{1mm}
\noindent \textbf{(2) Region Proposal Network (RPN):} Detects ROI (regions of interest) along with a confidence score, from the multi-scale feature maps generated by the backbone network. A fixed-size kernel is used for region pooling. The regions detected by the RPN are called \textit{proposal boxes}.

\vspace{1mm}
\noindent \textbf{(3) ROI Heads:} The input to the box head comprises 
(i)~the feature maps generated by a Fully Connected Network (FCN), 
(ii)~the \textit{proposed boxes} which come from the RPN. These are 1,000 boxes with their predicted labels. Box head uses the bounding boxes proposed by the RPN to crop and prepare the feature maps. 
(iii)~ground truth bounding boxes from the annotated training datasets. 
The ROI pooling uses the proposed boxes detected by RPN, crops the rectangular areas of the feature maps, and feeds them into the head networks. Using Box head and mask head together in Faster-RCNN network, inspired by He et al.~\cite{mask_rcnn} improves the 
 overall performance. 
 
During training, the box head makes use of the ground truth boxes to accelerate the training. The mask head provides the final predicted bounding boxes and  confidence scores during the training. At the time of inference the head network uses non-maximum suppression (NMS) algorithm to remove the overlapping boxes and selects the top-k results as the predicted output based on thresholds on their confidence score and intersection over union (IOU).

\begin{figure}[tb]
\centering
\includegraphics[width=0.95\textwidth, height=4.6cm]{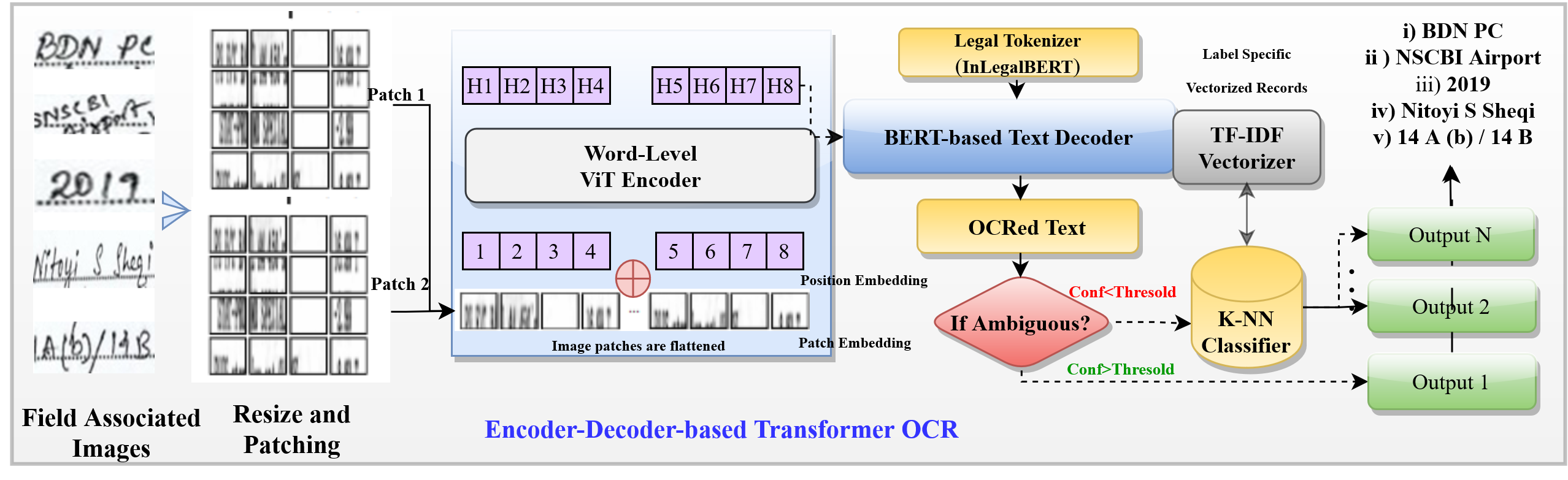}
\caption{TrOCR architecture with custom enhancements. The Text decoder uses a domain-specific InLegalBert~\cite{inlegalbert} based tokenizer. OCR predictions go for post-correction if the confidence score is less than the threshold. We convert the OCR prediction into a TF-IDF vector and search in the domain-specific field database to find the Nearest Match.} \label{fig:ocr-arch}
\label{fig:trocr-arch}
\vspace{-5mm}
\end{figure}

\subsection{The TrOCR architecture}

Once the localized images are generated for a target field (e.g., complainant name) by Faster-RCNN, the image patches are then flattened and sent to the Vision Transformer (ViT) based encoder model. 
We use TrOCR~\cite{trocr}  as the backbone model for our finetuning (see Figure~\ref{fig:trocr-arch}). TrOCR~\cite{trocr}  is a Transformer-based OCR model which consists of a pretrained vision Transformer encoder  and a pretrained text decoder. 
The ViT encoder is trained on the IAM handwritten dataset, which we fine-tune on our FIR dataset. 
We use the output patches from the Faster-RCNN network as input to the ViT encoder, and fine-tune it to generate features. As we are providing the raw image patches received from Faster-RCNN into the ViT encoder, we did not apply any pre-processing or layout enhancement technique to improve the quality of the localised images. 
On the contrary, we put the noisy localised images cropped from the \textit{form fields} directly, which learns to suppress noise features by training.

We also replace the default text decoder (RoBERTa) with the Indian legal-domain specific BERT based text decoder InLegalBERT~\cite{inlegalbert} as shown in Fig.~\ref{fig:ocr-arch}. 
InLegalBert~\cite{inlegalbert} is pre-trained with a huge corpus of about 5.4 million Indian Legal documents, including court judgements of the Indian Supreme Court and other higher courts of India, and various Central Government Acts. 
%A total number of 5.4 M documents are used in pre-training of this model.

To recognize characters in the cropped image patches, the images are first resized into square boxes of size 384 × 384 pixels and then flattened into a sequence of patches, which are then encoded by ViT into high-level representations and decoded by InLegalBERT into corresponding characters step-by-step.

We evaluate and penalise the model based on the Character Error Rate (CER). CER calculation is based on the concept of Levenshtein distance, where we count the minimum number of character-level operations required to transform the ground truth text into the predicted OCR output. CER is computed as $CER = ( S + D + I ) / N$
where $S$ is the number of substitutions, $D$ is the number of deletions, $I$ is the number of Insertions, and $N$ is the number of characters in the reference text.

\subsection{KNN-based OCR Correction}

\begin{figure}[tb]
\centering
\includegraphics[width=0.9\textwidth, height=5cm]{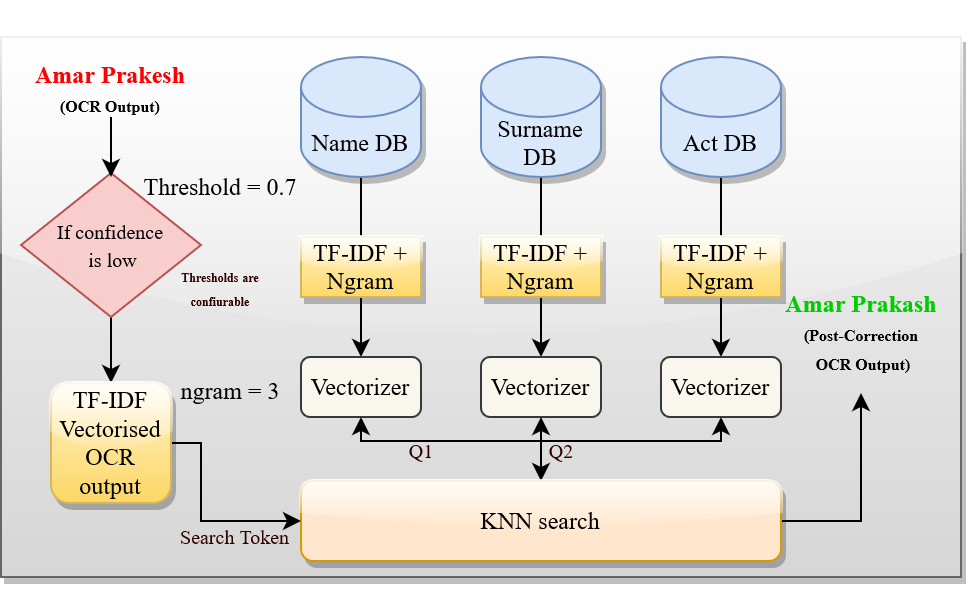}
\caption{Term Frequency and Inverse Document frequency (TF-IDF) Vectorizer based K-Nearest Neighbour model for post-correction on OCR output} \label{fig:knn-post-correct}
\end{figure}

For each predicted word from OCR, if the confidence score is less than a threshold 0.7, we consider the OCR output to be ambiguous for that particular word. 
In such cases, the predicted word goes through a post-correction step which we describe now (see Figure~\ref{fig:knn-post-correct}). 

\begin{table}[tb]
\caption{Excerpts from field-specific databases used to prepare TF-IDF vectorized records for KNN search. All databases contain India-specific entries.}
\label{tab:field-specific-databases}
\centering
\setlength{\tabcolsep}{4pt}
\begin{tabular}{ |c|c|c|c| } 
\hline
Names & Surnames & Police Stations & Statutes / Acts \\
 \hline \hline
Anamul & Haque & Baguiati & IPC (Indian Penal Code) \\ 
\hline
Shyam & Das & Airport & D.M. Act (Disaster Management Act) \\
\hline
Barnali & Pramanik & Newtown & D.C. Act (Drug and Cosmetics Act) \\ 
\hline
Rasida & Begam & Saltlake & NDPS Act \\ 
\hline
\end{tabular}
\end{table}

For each target field, we create a database of relevant values and terms (which could be written in the field) from various sources available on the Web. Table~\ref{tab:field-specific-databases} shows a very small subset of some of the field-specific databases such as Indian names, Indian surnames, Indian statutes (Acts and Sections), etc.
We converted each database into a set of TF-IDF vectors (see  Figure~\ref{fig:knn-post-correct}). 
Here TF-IDF stands for Term Frequency times Inverse Document Frequency.
The TF-IDF scores are computed using n-grams of groups of letters. In our work we used $n = 3$ (trigrams) for generating the TF-IDF vectors for OCR predicted words as well as for the entities in the databases.

For a given OCR output, based on the associated field name which is already available from the field classification by Faster-RCNN, we used the K-Nearest Neighbour (KNN) classifier to select the appropriate vectorized database. 
%During the inference phase, the pickle file associated with the classified field name is  selected for K-Nearest Neighbour search.
KNN returns best matches with a confidence score based on the distance between the search vector (OCR output) and the vectors in the chosen database. If the confidence score returned by KNN is greater than $0.9$, then the OCR predicted word gets replaced with the word predicted by the K-Nearest Neighbour search.

\section{Experimental settings}

We ran all experiments on a Tesla T4 GPU with CUDA version 11.2. We used CUDA enabled Torch framework 1.8.0. 

In the first stage of the TransDocAnalyser framework, we trained the Faster RCNN from scratch using the annotated dataset (the training set). Table~\ref{tab:faster-rcnn-parameters} shows the settings used for training the Faster-RCNN model. 
Prior to the training, input images are resized in 1180 $\times$ 740. For memory optimization, we run the model in two steps, first for 1500 iteration and then for 1000 iteration on the stored model. We tried batch sizes (BS) of 16, 32 and 64, and finalized BS as 64 because of the improvement in performance and training time.
We used the trained model Faster-RCNN model to detect and crop out the bounding boxes of each label from the original document (as shown in Fig.~\ref{fig:sample-entities}) and created our dataset to fine-tune the ViT encoder. 

We also created a metadata file mapping each cropped image (as shown in Fig.~\ref{fig:sample-entities}) with 
its corresponding text as described in~\cite{iam_dataset} to fine-tune the decoder. 

\begin{table}[tb]
\caption{Faster-RCNN model training parameters}
\label{tab:faster-rcnn-parameters}
\centering
\begin{tabular}{ |c|c|c|c|c|c|c| } 
\hline
Base Model & Base Weights & Learning Rate & Epoch \#  & \# of Class & IMS/batch & Image Size\\
\hline
ResNet 50 &  Mask RCNN & 0.00025 & 2500 & 4 & 4 & 1180 $\times$ 740\\
 \hline
\end{tabular}
\end{table}

\begin{table}[tb]
\caption{Tranformer OCR (TrOCR) parameters used for model fine-tuning}
\label{tab:trocr-parameters}
\centering
\begin{tabular}{ |l|l|l|l|l|l|l| } 
 \hline
Feature Extractor & Tokenizer & Max Len & N-gram & Penalty & \# of Beam & Optimizer \\
 \hline
google-vit-patch16-384 & InLegalBERT & 32 & 3 & 2.0 & 4 & AdamW \\
 \hline
\end{tabular}
\vspace{-5mm}
\end{table}

Table~\ref{tab:trocr-parameters} shows the parameter settings used for fine-tuning the TrOCR model.
Image patches are resized to 384 $\times$ 384 dimension to fine-tune ViT encoder. In the TrOCR model configuration, we replaced the tokenizer and decoder settings based on InLegalBert.
We tried with batch size (BS) of 2, 4, 8, 16, 32, 64, and BS = 8 provided the best result on the validation set. We fine-tuned the Encoder and Decoder of the OCR for 40 epochs and obtained the final results.

The KNN-based OCR correction module used n-grams with $n = 1,2,3,4$ to generate the TF-IDF vectors of the field-specific databases.
Using $n = 3$ (trigrams) and KNN with $K = 1$ provided the best results.

%To increase the volume of the dataset, we prepared the dataset in two iteration. At first, we manually annotated 50\% of the images and train all the models and ran inference on the rest 50\% images of the dataset. Annotators then reviewed the performance of the models and corrected the results by annotating those images which then used to finally train the models and the all configuration, results and metrics we presented from the final experiment.

%%%%%%%%%%%%%%%%%%%%%%

\begin{table}[tb]
\caption{Performance of field labelling on the FIR dataset (validation set and test set). Re: Recall, Pr: Precision, F1: F1-score, mAP: mean average precision.}
\label{tab:faster_rcnn_metrics}
\setlength{\tabcolsep}{4pt}
\centering
\begin{tabular}{|l|l|l|l|l|l|}
\hline
Results on dataset & Target field & \multicolumn{4}{|l|}{Faster R-CNN} \\
\cline { 3 - 6 }
 &  & $\mathbf{Re} \uparrow$ & $\mathbf{Pr} \uparrow$ & $\mathbf{F1} \uparrow$ & $\mathbf{m A P} \uparrow$ \\
\hline
Validation & Year & $0.98$ &  $0.96$ & $0.97$ & $0.97$ \\
 & Statute & $0.85$ & $0.82$ & $0.83$ & $0.84$ \\
 & Police Station & $0.96$ & $0.90$ & $0.93$ & $0.93$ \\
 & Complainant Name & $0.84$ & $0.76$ & $0.80$ & $0.77$ \\
 
\hline
Test & Year & $0.97$  & $0.96$ & $0.97$ & $0.96$  \\
 & Statute & $0.84$  & $0.87$ & $0.86$ & $0.80$ \\
 & Police Station & $0.93$  & $0.88$ & $0.91$ & $0.91$ \\
 & Complainant Name & $0.80$  & $0.81$ & $0.81$ & $0.74$ \\
\hline
\end{tabular}

\end{table}

\section{Results}

In this section, we present the results of the proposed framework TransDocAnalyser in three
stages -- 
(i)~The performance of Faster-RCNN on localization and labelling of the target fields (Table~\ref{tab:faster_rcnn_metrics});
(ii) Sample of OCR results with Confidence Scores (Table~\ref{tab:ocr_errors}); and (iii) Comparison of the performance of the proposed framework with existing OCR methods (Table~\ref{tab:ocr_benchmarking}).

Table~\ref{tab:faster_rcnn_metrics} shows the results of field label detection using Faster-RCNN on both test and validation sets of the FIR dataset. The performance is reported in terms of Recall (Re), Precision (Pr), F1 (harmonic mean of Recall and Precision) and mean Average Precision (mAP). 
For the localization and labelling, a prediction is considered correct if both the IOU (with the ground truth) and the confidence threshold are higher than $0.5$. 
The results show that our model is performing well, with the best and worst results for the fields `Year' (F1 = 0.97) and `Name' (F1 = 0.8) respectively. This variation in the results is intuitive, since names have a lot more variation than the year.

\begin{figure}[tb]
\centering
\includegraphics[width=\textwidth]{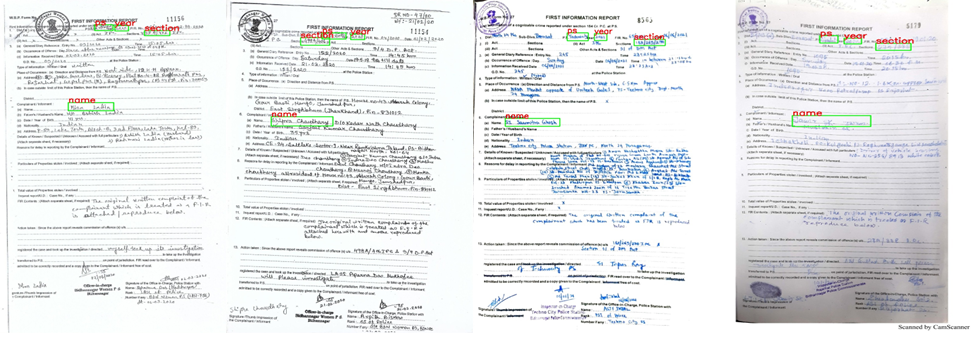}
\caption{Examples of localization and labelling of target fields by Faster-RCNN. The predicted bounding boxes are highlighted in green on the images. The associated class labels are  highlighted in red.} 
\label{fig:faster-rcnn-output-examples}
\end{figure}

Figure~\ref{fig:faster-rcnn-output-examples} shows examples of outputs of Faster-RCNN on some documents from the test set of the FIR dataset. 
The predicted bounding boxes are highlighted in green rectangles, and the predicted class names are marked in red on top of each bounding box.

\begin{table}[tb]
\caption{Finetuned (TrOCR) predictions on the generated image patches shown below}
\label{tab:ocr_errors}
\centering
\begin{tabular}{ |c|c|c| } 
\hline
{\bfseries Image Patches} & {\bfseries OCR Results} &  {\bfseries Confidence Score}\\
\hline
\includegraphics[width=3cm, height=0.5cm]{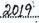} & 2019 & 0.89\\
\hline
\includegraphics[width=5.5cm, height=0.5cm]{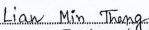} & Lian Min Thang & 0.77\\
\hline
\includegraphics[width=5cm, height=0.5cm]{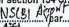} & Nscbi Airport & 0.79\\
\hline
\includegraphics[width=5.5cm, height=0.5cm]{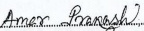} & Amar \textcolor{red}{ Prakesh}  & \textcolor{red}{0.63}\\
\hline
\includegraphics[width=3cm, height=0.5cm]{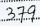} & 379 & 0.96\\
\hline
\end{tabular}
\vspace{-5mm}
\end{table}

The output of Faster-RCNN provides bounding boxes and field names for each image, using which image patches are generated and sent to the Encoder-Decoder architecture.
Table~\ref{tab:ocr_errors} shows some examples of image patches and the finetuned TrOCR predictions for those image patches.
It is seen that the name ``Amar Prakash'' is predicted as `Amar Prakesh'' with confidence score below a  threshold of 0.7 (which was decided empirically). As the prediction confidence is below the threshold, this output goes to the post-correction method proposed in this work.

Table~\ref{tab:ocr_benchmarking} compares the final performance of our proposed
framework TransDocAnalyser, and compares our model with Google-Tesseract and Microsoft-TrOCR for handwritten recognition on proposed FIR dataset.\footnote{We initially compared Tesseract with TrOCR-Base, and found TrOCR to perform much better. Hence subsequent experiments were done with TrOCR only.}
The performances are reported in terms of Character Error Rate (CER), Word Error Rate (WER), and BLEU scores~\cite{bleu_score}. 
Lower values of CER and WER indicate better performance, while higher BLEU scores are better. 

\begin{table}[tb]
\caption{Benchmarking state-of-the-art TrOCR and our proposed framework TransDocAnalyser on the FIR dataset (best values in boldface)}
\label{tab:ocr_benchmarking}
\centering
\begin{tabular}{|c c | c c c|}
\hline
\bfseries OCR models & \bfseries Target Field & \multicolumn{3}{|l|}{\bfseries Evaluation Metrics} \\
\cline {3 - 5}

 &  & $\mathbf{CER} \downarrow$  & $\mathbf{WER} \downarrow$ & $\mathbf{BLEU} \uparrow $\\
\hline
Tesseract-OCR & Year & $0.78$ &  $0.75$ & $0.14$ \\
 & Statute & $0.89 $ &  $0.83$ & $0.12$ \\
 & Police Station & $0.91$ &  $0.89$ & $0.10$ \\
 & Complainant Name & $0.96$ &  $0.87$ & $0.9$ \\
 
\hline
TrOCR-Base & Year & $0.38$ &  $0.32$ & $0.72$ \\
 & Statute & $0.42 $ &  $0.38$ & $0.68$ \\
 & Police Station & $0.50$ &  $0.44$ & $0.62$ \\
 & Complainant Name & $0.62$ &  $0.56$ & $0.56$ \\
 
\hline
TrOCR-Large & Year & $0.33$ &  $0.32$ & $0.75$ \\
 & Statute & $0.34 $ &  $0.33$ & $0.73$ \\
 & Police Station & $0.36$ &  $0.38$ & $0.65$ \\
 & Complainant Name & $0.51$ &  $0.50$ & $0.57$ \\

\hline
TrOCR-\textbf{InLegalBert} & Year & $0.17$ &  $0.17$ & $0.84$ \\
 & Statute & $0.19 $ &  $0.21$ & $0.92$ \\
 & Police Station & $0.31$ &  $0.26$ & $0.78$ \\
 & Complainant Name & $0.45 $ &  $0.39$ & $0.72$ \\
 
\hline
\textbf{TransDocAnalyser (proposed)} & Year & $\mathbf{0.09}$ &  $\mathbf{0.02}$ & $\mathbf{0.96}$ \\
 & Statute & $\mathbf{0.11} $ &  $\mathbf{0.10}$ & $\mathbf{0.93}$ \\
 & Police Station & $\mathbf{0.18}$ &  $\mathbf{0.20}$ & $\mathbf{0.83}$ \\
 & Complainant Name & $\mathbf{0.24} $ &  $\mathbf{0.21}$ & $\mathbf{0.78}$ \\

\hline
\end{tabular}
\vspace{-5mm}
\end{table}

%We perform bench-marking combining each component of this framework and the results are shown in the Table 8. 
We achieve state-of-the-art results using the proposed TransDocAnalyser framework which outperforms the other models with quite a good margin (see Table~\ref{tab:ocr_benchmarking}). %Tesseract-OCR performed worse on detection of the handwritten characters. 
While the TrOCR + InLegalBert model also performed well, 
%with 0.232 character error rate (CER) and 0.76 Bleu score whereas 
our proposed framework TransDocAnalyser (consisting of vision transformer-based encoder, InLegalBert tokenizer and KNN-based post-correction) achieved the best results across all the four target fields of the FIR dataset.

\section{Conclusion}

In this work, we (i)~developed the first dataset for semi-structured handwritten document analysis in the legal domain, and (ii)~proposed a novel framework for offline analysis of semi-structured handwritten documents in a particular domain. 
Our proposed TransDocAnalyser framework including Faster-RCNN, TrOCR, a domain-specific language model/tokenizer, and KNN-based post-correction outperformed  existing OCRs.

%To our knowledge, this is the first work considering the challenges related to Indian legal document analysis.
We hope that the FIR dataset developed in this work will enable further research on legal document analysis which is gaining importance world-wide and specially in developing countries.
We also believe that the TransDocAnalyser framework can be easily extended to semi-structured handwritten document analysis in other domains as well, with a little fine-tuning.

% Moreover, the improved version with ambiguity solver using TFIDF + KNN model shows that it achieved state-of-the-art results on semi-structured offline handwritten FIR documents.

%\section{Acknowledgement}
\vspace{4mm}
\noindent \textbf{Acknowledgement:}
This work is partially supported by research grants from Wipro Limited (\url{www.wipro.com}) and IIT Jodhpur (\url{www.iitj.ac.in}).


%

\end{document}